\documentclass[runningheads]{llncs}
\usepackage{graphicx}
\usepackage{amsmath,amssymb} 
\usepackage{color}
\usepackage{amsmath}
\usepackage[width=122mm,left=12mm,paperwidth=146mm,height=193mm,top=12mm,paperheight=217mm]{geometry}

\usepackage{verbatim} 

\let\emptyset\varnothing
\usepackage{dcolumn}
\usepackage{multirow} 
\usepackage{slashbox} 
\usepackage{tabularx} 
\usepackage[linesnumbered,ruled,vlined]{algorithm2e}
\usepackage{pifont}

\usepackage{graphicx}
\usepackage{caption}
\usepackage{subcaption}
\DeclareGraphicsExtensions{.pdf,.png,.jpg}

\graphicspath{ {./} }

\begin{document}
\pagestyle{headings}
\mainmatter
\def\ECCV14SubNumber{}  

\title{
Joint Energy-based Detection and Classification of Multilingual Text Lines} 

\titlerunning{}

\authorrunning{}

\author{ 
	Igor Milevskiy\hspace{3ex}     Yuri  Boykov \\Computer Science Department \\University of Western Ontario, Canada \\ {\tt\small io.milewski@gmail.com, \hspace{1ex} yuri@csd.uwo.ca}}

\institute{}
\maketitle

\begin{abstract}
This paper proposes a new hierarchical MDL-based model for a joint detection and classification of multilingual text lines in images taken by hand-held cameras.
The majority of related text detection methods assume alphabet-based writing in a single language, e.g. in Latin. They use simple clustering heuristics specific to such texts: 
proximity between letters within one line, larger distance between separate lines, etc. 
We are interested in a significantly more ambiguous problem where images combine alphabet and logographic characters from multiple languages and typographic rules vary a lot (e.g. English, Korean, and Chinese). Complexity of detecting and classifying text lines in multiple languages calls for a more principled approach based on information-theoretic principles. Our new 
MDL model includes data costs combining geometric errors with classification likelihoods and
a hierarchical sparsity term based on {\em label costs}. This energy model can be efficiently minimized by {\em fusion moves}. We demonstrate robustness of the proposed algorithm on a large new
database of multilingual text images collected in the public transit system of Seoul.
\end{abstract}

\section{Introduction}
Over the centuries, people in different parts of the world developed various writing systems. 
The most common writing systems are Latin, Cyrillic, Arabic, Logographic Chineese characters, such as Hanzi, Haja and Kanji (in use in China, Korea, and  Japan respectively), and alphabet based Hangul (Korea), Kitakana  and Hiragana (both Japan).
Most people comprehend text in one or at most two writing systems. Thus, it is extremely helpful to dedicate reading functions to a computing device. Significant progress in text detection and recognition was made in the scope of printed documents. 
Recently, researchers focus on more challenging tasks, such as text detection in video, web, and camera captured images 	\cite{CVPR1,CVPR3,CVPR4,CVPR2}, which is also the main focus of this work.

For a single line of text (e.g. in on-line handwriting recognition) or when multiple lines are easy to detect 
(e.g. in printed documents) the problem of text recognition can be solved with well established techniques such as the Hidden Markov Model (HMM) and Recurrent Neural Networks (RNN) \cite{PAMI_BLSTM}.
The input for these algorithms is an image segment
corresponding to one line of text. In particular, this segment
does not need to be split into isolated characters.

\begin{figure}[t]
\begin{center} 
   \includegraphics[width=0.8\linewidth]{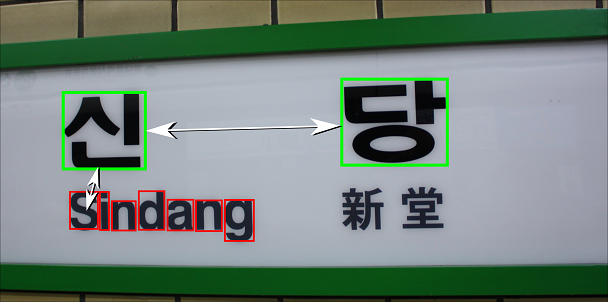}
\end{center}
   \caption{Signboard in three languages. Korean and English text lines are marked with boxes. Two Korean character form one word. There is a significant gap between them, which makes the text detection problem challenging.}
\label{fig:kr_distant} 
\end{figure}

\begin{figure}[t]
\begin{center}
   \includegraphics[width=1.0\linewidth]{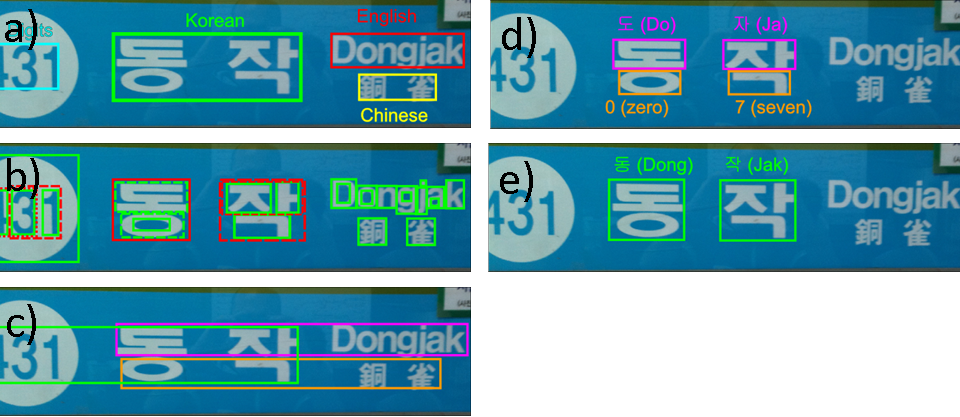}
\end{center}
\caption{ Examples of ambiguities in multilingual text line detection. a) Ground truth with Korean, English, Chinese characters and digits. 
b) Detected blobs. Green boxes come from connected component analysis. Red boxes are placed using heuristics to cover potential Korean characters  c) Minimization of  geometric errors alone leads to incorrect grouping. d-e) Classification ambiguities: d) two Korean characters on top, two digits on the bottom. e) two Korean characters are seen. Both d) and e) are equally good for the classifier.}  
\label{fig:combined}
\end{figure}

In the context of camera captured images, e.g. Fig.\ref{fig:kr_distant}, detection of text lines is
an additional challenging problem that must be solved before image segments containing separate 
lines can be sent to the recognizer. Accurate partitioning of multiple text lines is critical for the success 
of the recognizer. This paper focuses specifically on the task of individual text line recognition, 
which is particularly challenging in presence of multiple languages, see Fig.\ref{fig:combined}(a).

Most of the prior work on text line detection deals with scripts in Latin characters printed according to one standard set of typographical rules. In Asia, however, signboards on streets, in subways stations, railway platforms and bus terminals can have text in two or more languages at once. We aim to detect multiple lines of text from an image, e.g. Fig.\ref{fig:kr_distant}, that has characters in several languages arranged according to different typographical regulations.

For example, English typography includes letters in upper and lower cases, some characters also have descenders and ascenders. English letters can cross the mean and the baseline. One common convention is that English letters sit close to each other within each line of text, which makes text line detection relatively easy.

In contrast, Korean typographical rules make it harder to detect lines of text for two reasons. The first one is illustrated in Fig.\ref{fig:kr_distant} where two green boxes mark Korean characters (or syllabuses) belonging to the same word. The gap between these characters is significantly greater than the distance between the first Korean character and the English line below it. Such large inter-character gaps are very common in Korean typography. 

The second difficulty is that multiple Korean letters can sit on top of each other within a single text line. For example,
each green box in Fig.\ref{fig:kr_distant} corresponds to a
syllabus containing 3 letters.  As shown in Fig.\ref{fig:combined}(c and d), it is easy to incorrectly group individual Korean letters in the same syllabus 
into multiple lines. Fig.\ref{fig:combined}(a) shows the correct segmentation.

Similarly to Korean, Chinese characters often disperse into pieces. Thus, it could be hard to determine if one piece (a detected blob) corresponds to a character or a part of a character. Another similarity between Chinese and Korean characters distinguishing them from English letters is their fairly consistent aspect ratio, typically around one.



Multilingual text detection in camera-captured images represents a particular case of scene understanding over a small set of object classes (languages). 
The general scene understanding problem (e.g. on PASCAL data) is commonly solved 
by partitioning an image into segments of different classes, e.g.\cite{Ladicky:eccv10}. However, most methods do not separate distinct object instances within one class\footnote{Some methods use object detectors to count object instances \cite{Ladicky:eccv10_Instances}}\@. The technical challenge of our multilingual text detection problem is that we have to segment an image into individual instances of objects (text lines) from multiple categories (languages).
We formulate this as a multi-model fitting problem based on {\em label cost} regularization \cite{CVPR_Andrew_Hosam_LabelCost}. The original image is represented by a set of detected blobs, which are assigned different models/labels. Each model (text line) is described by geometric parameters\footnote{Due to perspective distortion we represent {\em base-} and {\em mean lines}.} and an additional category parameter - language (Fig. \ref{fig:geometry}). This category defines how the data fitting errors are computed accounting for typographical 
differences between the languages.

There is a lot of prior work on geometric model fitting in vision. RANSAC is the most common method 
for data supporting a single model. It is known for its robustness to outliers. 
Multi-model problems are often addressed by procedurally-defined clustering heuristics
greedily maximizing inliers, e.g. Mutli-RANSAC \cite{MULTIRANSAC}, Hough space mode selection, or 
{\em j-linkeage} \cite{Toldo:JLINK}. These methods often fail on difficult examples with weak data in the presence of much noise or outliers \cite{IJCV_Hossam}.
The practical challenges of multilingual multi-line text detection motivates more powerful approaches
based on optimization of a clearly defined objective function, in particular, MDL criteria \cite{CVPR_Andrew_Hosam_LabelCost,TORR}. Many previous MDL methods fit geometric models of the same class, e.g. lines.
Some techniques fit a hierarchy of geometric models like 
{\em points$\rightarrow$lines$\rightarrow$cubes} \cite{SongChunZhu:AerialView} or {\em points$\rightarrow$lines$\rightarrow$vanishing points} \cite{Delong:nips12}. 
We approach the multilingual text line detection problem by fitting geometric models (lines) from 
independent appearance-based categories (language) that can be interpreted as a hierarchy
{\em blobs$\rightarrow$lines$\rightarrow$languages}. 

Our contributions are summarized below:
\begin{itemize}

\item  We propose a new challenging application for computer vision:  multilingual text line detection over languages with different typography rules. 
Our database of camera-captured multilingual text images with ground truth 
(500 images) will be made publicly available. 

\item We propose a novel hierarchical MDL energy for multilingual text detection \eqref{eq:FusionFULLFUSION}. 
Our sparsity-based (label cost) regularization is applicable to a wider range of typographies compared to 
smoothness-based approaches \cite{JOURNAL1,ICDAR5} assuming proximity between letters. 
Our energy can be efficiently optimized by fusion moves \cite{Delong:nips12}.

\item In contrast to many standard techniques for scene understanding, our method simultaneously segments 
an image into multiple classes (languages) and into individual object instances (text lines) within each class by observing geometric errors and classification likelihood.

\item Unlike previous text detection algorithms assuming a single language
with no grouping ambiguities as in Fig.\ref{fig:combined}, 
we solve a much harder clustering problem that can not be addressed with
standard simple heuristics. Instead, we have to formulate and optimize an information theoretic criteria to resolve all ambiguities in the proposed ill-posed problem. Our MDL approach is closely related to semantic segmentation.
It has Bayesian interpretations \cite{LECLRERC_89,PAMI_96:Zhu_Yuille,TORR,IJCV:Label_cost}.
\end{itemize}

The rest of the paper is organized as follows. Sec.\ref{sec:related_work} discussed related prior work on text detection. 
Sec.\ref{sec:algorithm} presents our new  hierarchical MDL formulation of the multilingual generalization of this problem.
Our camera-captured text image database and experimental evaluation of our algorithm is presented in Sec.\ref{sec:eval}.

\section{Related work}  \label{sec:related_work}
Interest in text detection has increased since 2003 when the ICDAR database and competition was formed. After that, ICDAR 2011, Street View Text(SVT) and private publicly unavailable databases were collected. All mentioned databases include text in Latin script. A number of text detection algorithms were proposed and evaluated on these image collections. 

In general, a text detection algorithm combines the following parts: text candidate detection, text candidate filtering and line fitting. 
For a given input image the algorithm produces a set of rectangles either horizon-oriented or rotated.

There are three major groups of text candidate detection methods: sliding window \cite{CVPR3,JOURNAL1,ICDAR9,ECCV3,ICDAR4,ICDAR2}, edge based  \cite{ICDAR5,ICCV2,ICDAR7}, and color based \cite{CVPR4,ACCV2} algorithms.

Sliding window algorithms, originally proposed for face detection, denote an exhaustive search. Features are computed for each position and scale of the sliding window. 

Edge based methods retrieve an edge map (Sobel, Canny, Laplace) and then perform connected component (CC) analysis and outputs blobs. Moreover, Stroke width transform (SWT) \cite{CVPR1,CVPR2} also aims to find blobs that have consistent width of stroke.
Color based methods, such as MSER and ER (inspired by MSER) assume that a text character's color is homogeneous. MSER was used by the ``ICDAR 2011 robust reading competition'' winner.

After text candidates are detected non-text blobs must be filtered out. The decision whether a blob represents text is done by classification. Popular classifiers are: support vector machine (SVM) \cite{ICCV2,ACCV2,ICCV4} , AdaBoost \cite{ICDAR3,ICDAR1} or their cascades. Popular features for classification are:
color based (histogram of intensities, moment of intensity)
edge based (histogram of oriented gradients (HOG), Gabor filter) geometric features (width, height, aspect ration, number of holes, convex hull, area of background/foreground).

Single text blobs must then be aggregated into text lines. Older approaches are based on the Hough transform algorithm \cite{JOURNAL2,JOURNAL98JAPAN}.
More recent algorithms combine neighbours of blobs into pairs and then fulfill clustering in N-dimensional space, where the following dimensions are in use: stroke width, orientation of a pair, and geometric size of blobs.

Text candidate filtering and text line detection could be done consequentially with complex approaches based on a Markov random field (MRF) \cite{CVPR3,ICDAR4} or a Conditional Random Field (CRF) \cite{ICDAR_PAN} and algorithms based on minimal spanning trees.

In \cite{JOURNAL2,ICCV3} the authors assume that text is located on a signboard. They try to locate physical edges and corners of that signboard and perform signboard rectification.

\section{Algorithm}   \label{sec:algorithm}
\begin{figure}[t]
\begin{center}
   \includegraphics[width=1.0\linewidth]{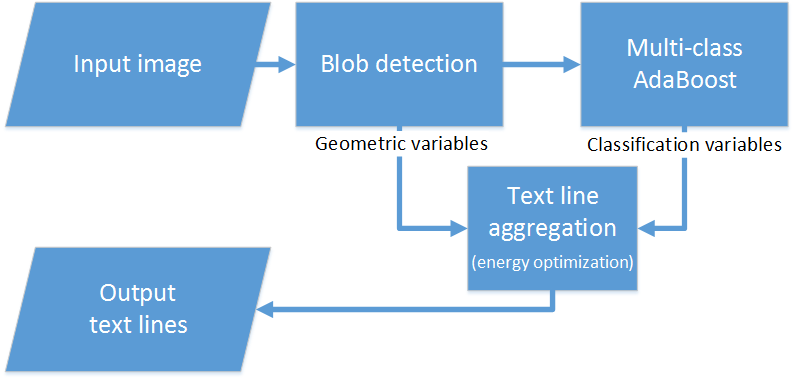}
\end{center}
   \caption{Energy-based algorithm for joint text line detection and classification. }
\label{fig:workflow}
\end{figure}

The general work flow of our algorithm is shown in Fig.~\ref{fig:workflow}. First, in the input image blobs or text candidates are detected. Next, in addition to geometric properties each blob gets an likelihood to belong to one of the five categories (English, Korean, Chinese, Digit, Non-text). The likelihood is provided by an AdaBoost classifier. Finally, energy-based algorithm groups blobs into text lines. This can be seen as text candidate clustering in multidimensional space of geometric and classification values.

Let us recall the challenges of the multilingual text detection problem:
\begin{itemize}
\item The number of languages and the number of text lines is not known.
\item A text candidate can occupy a whole text string's height or a part of that sting.
In other words there is an ambiguity of grouping text candidates of an Asian script (See Fig \ref{fig:combined}). Therefore, a sequential classification-aggregation algorithm i.e. \cite{ICDAR_PAN} is not applicable to our problem.
\item Smoothness term, standard to MRF, can not be used since it makes sense only for European scripts (See Fig \ref{fig:kr_distant}).
\end{itemize}

The MDL principle suggest that the solution of the problem is the smallest set of text lines which fits text candidates with the smallest geometric and classification errors. The solution must have text lines in only a few languages.

Our joint text detection and classification approach is based on the hierarchical energy minimization:
\begin{align}
 E(\boldsymbol{l},\boldsymbol{\theta})  = \sum\limits_{i\in \mathcal{I}}{D_i(l_i | \boldsymbol{\theta})} + \sum\limits_{j\in \mathcal{L}}{C_j \cdot [ \exists{l_i = j} ] } + \sum\limits_{v\in \mathcal{V}}{C_v \cdot [ \exists{v_{l_i} = v} ] },
\label{eq:FusionFULLFUSION} 
\end{align}
 where $ \mathcal{I} $ is the set of all text candidates. Text candidate detection is described in Subsec. \ref{Blob detection}. $ \mathcal{L} $ is the space of all possible text line models. $ \mathcal{V} $ is the space of all possible languages. The vector $ \boldsymbol{\theta} $ contains text line model parameters:
 \begin{align}
 \boldsymbol{\theta} = \{\theta_j | j \in \mathcal{L} \} .
 \label{eq:model_param_definition} 
 \end{align}
 Text line model parameters $\theta_j$ are a language mark and parameters of base and mean lines (See Fig. \ref{fig:geometry}). The lines of a model are not necessarily parallel. The model can describe perspectively distorted or arbitrary rotated text lines in either English, Chineese or Korean.
 The vector $ \boldsymbol{l}$ contains
  indexes of text line models assigned to each text candidate $ i $ :
  \begin{align}
  \boldsymbol{l} = \{l_i \in \mathcal{L} | i \in \mathcal{I}\} .
  \label{eq:labeling_definition} 
  \end{align} 
$ D_i( j | \boldsymbol{\theta})$, the data term, tells how well a particular text candidate $ i $ fits text line model $ j $:
 \begin{align}
 D_i(j | \boldsymbol{\theta}) =\overbrace{ -ln Pr(i | lang_j)}^{classification} + \overbrace{dist_{lang(j)}(i,lines_j)}^{geometry}.
 \label{eq:Error function geom and likelihood} 
 \end{align}
 The data term consist of two values: the classification likelihood (Subsec. \ref{classification}) and the geometric error (Subsec. \ref{geometry}). The geometric error is scaled according to the language of the model. The outlier cost $D_i(\emptyset) = const$ is paid if the blob is not associated with any text line model. Outlier model works like a non-text blob collector.

$ C_j $ penalizes number of text line models in the solution. The penalty $ C_j $ is counted if there is at least one data point $ i $ assigned to the text line model $ j $. $ C_v $ penalizes number of languages in the solution. The penalty $ C_v $ is counted if there is at least one data point $ i $ assigned to the language $ v $.

The solution of the optimization problem \eqref{eq:FusionFULLFUSION} is the set of text line model parameters $\boldsymbol{\theta}$, and labeling $\boldsymbol{l}$. A labeling maps text candidates into the text lines.

The energy \eqref{eq:FusionFULLFUSION} describes the following hierarchical levels: {\em blobs} $\rightarrow$ {\em text lines} $\rightarrow$ {\em languages}.

\subsection{Blob detection and proposal}
\label{Blob detection}


We use an edge based blob detector to produce text candidates.
At first, an input image is converted to gray-scale and downsized for performance reasons. Next, horizontal and vertical Sobel filters are applied. Two resulting images are combined. The result represents an edge map. Finally, connected component analysis run resulting character candidates (blobs).

Unlike a color based blob algorithm, an edge based method does not have the foreground-background ambiguity when algorithm must be run twice for dark text on bright background or vice versa.

More complex blob detectors were considered. Chinese symbols are very compact. They have high density of strokes per unit of area. In such conditions, SWT fails to detect Chinese characters well. 

At this point there are Korean characters that are over-segmented (Fig. \ref{fig:combined} b). It is necessary to propose more text candidates to cover such characters. Two or three blobs produce a new blob if they stand close to each other, cover resulting area well and the resulting aspect ratio is close to one.

\subsection{Geometric errors} \label{geometry}
 
\begin{figure}[t]
\begin{center}
   \includegraphics[width=0.5\linewidth]{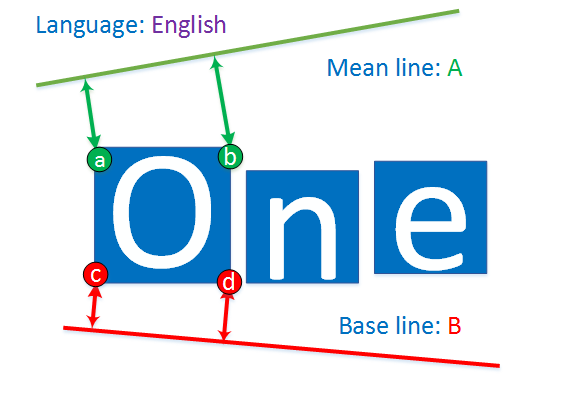}
\end{center}
   \caption{Geometric errors.}
\label{fig:geometry}
\end{figure}

A text candidate $i$ is more likely to be a part of a text line $j$ if it sits close to its base and mean lines. The geometric part of the data term \eqref{eq:Error function geom and likelihood} includes the sum of Euclidean distances between four corners of a text candidate's box and the base and the mean line of a model (Fig. \ref{fig:geometry})

\begin{align}
dist_{lang(j)}(i,lines_j) = \\ \nonumber
\frac{K_{lang(j)}} {Z_j} \cdot ( &||A_j - a_i|| +\\ \nonumber
&|| A_j - b_i||  +\\ \nonumber
&|| B_j - c_i|| +\\ \nonumber
&|| B_j - d_i|| ),
\end{align}
where $K_{lang(j)}$ is a language dependent scale. There are ascender and descender characters in English alphabet. In contrast to English, Korean and Chinese characters have roughly same height. We allow larger tolerance for geometric error between the text candidates and English lines, and smaller tolerance for geometric errors between the text candidates and Chinese or Korean text lines. $Z$ is a constant that depends on the height of text line $j$.

\subsection{Classifier and features}
 \label{classification}

AdaBoost is a machine learning algorithm that builds a strong classifier out of weak ones. AdaBoost have become a popular choice for text vs. not-text recognition. In our problem we use a multi-class AdaBoost, which is trained on the following categories: English, Korean, Chinese, Digit, Non-Text. Decision trees are commonly used as a weak classifier. It is possible to scale the classifier according to the complexity of the problem by adjusting the depth and the quantity of the trees.

A blob that is marked as ``text'' must cover a text line's height completely. If it covers two or more lines (case of under-segmentation), or only a portion of a text line's height (case of over-segmentation) then it is treated as non-text.

The following groups of features are used for AdaBoost classifier training: colour, gradient and geometry based features.

A bounding box that contains text has normally two distinctive colors  - one for foreground, one for background. Non-text blobs do not have such consistency. Color features are useful for text vs. non-text classification. Histogram of intensities descriptor is computed as in \cite{ICDAR3}. Text of a different language could be diversified according to stroke density and number of strokes in a particular direction. We use histogram of oriented gradients (HOG) and Gabor filter to describe such properties. Geometry based features are blob's width, height and width-to-height ratio.

\subsection{Energy optimization details}
Minimization of energy \eqref{eq:FusionFULLFUSION} is an NP-hard mixed optimization problem combining integer labeling variables and real-valued model parameters. Nevertheless, there are approximation algorithms
\cite{CVPR_Andrew_Hosam_LabelCost,IJCV_Hossam,Delong:nips12}, 
which guarantee some optimality bounds and, in practice, give good approximate 
solutions. 

We adopt {\em PEARL}, see Algorithm \ref{algo:PEARL}, introduced by Isack et al. \cite{IJCV_Hossam}. The first step produces a finite set of models by randomly sampling data points as in RANSAC. Then, block coordinate descend (BCD) procedure 
minimizes energy \eqref{eq:FusionFULLFUSION} iterating two steps: 
({\em AssignModels}) selecting 
an optimal labeling by optimizing over discrete variables $\boldsymbol{l}$ 
and ({\em RefitModels}) optimizing real-valued model parameters $\boldsymbol{\theta}$. 

\begin{algorithm}
\DontPrintSemicolon 
\SetKwInOut{Input}{Input}  
\Input{$data\_points$ {//} blobs, etc.\; 
}
\SetKwInOut{Output}{Output}  
\Output{$labeling$ {//} data\_points-to-model map  \; \\
$models$ {//} text lines 
}
$pool \gets RandomSampleModels(data\_points);$\;

\For{$i\leftarrow 0$ \KwTo $max\_iterations$} 
{ 
 $labeling   \gets AssignModels(pool,data\_points);$\;
 $pool   \gets RefitModels(pool,data\_points,labeling);$\;
}
$models   \gets pool$\;
\Return{$[models, labeling];$}\;
\caption{{\sc PEARL {}// energy-based model fitting algorithm} }
\label{algo:PEARL}
\end{algorithm}

We compute our initial set of models as follows. First, a neighbourhood map is build by Delaunay triangulation. Next, blobs that have a common edge produce a model. Unlike k-nearest neighbour, such an approach enables a pair of blobs to produce a text line model, regardless of the distance between the blobs.

{\em AssignModels} routine assigns each blob to a model (label) from the pool based on an optimal labeling with respect to hierarchical MDL 
energy \eqref{eq:FusionFULLFUSION} assuming fixed model parameters $\boldsymbol{\theta}$. The corresponding optimization algorithm is described in details below.
{\em RefitModels} step re-estimates optimal model parameters $\boldsymbol{\theta}$
assuming fixed labeling $\boldsymbol{l}$. In this case, only the
data term of energy \eqref{eq:FusionFULLFUSION} depends on parameters
$\boldsymbol{\theta}$. Thus, optimal geometric parameters for each model
should be independently computed for each set of inliers, that is, 
to blobs assigned the same model (label). The top and the bottom lines of a single model are refit by linear least squares (LS).
 
{\bf Labeling Optimization:} Procedure {\em AssignModels} computes an
optimal discrete labeling $\boldsymbol{l}$ for energy
\eqref{eq:FusionFULLFUSION} with fixed parameters
$\boldsymbol{\theta}$ using the standard {\em fusion move} or {\em optimized crossover} framework \cite{PAMI_10_Lempitsky,Delong:nips12}, as summarized in Algorithm \ref{algo:AssignModels}. The main idea of {\em fusion moves} is
to iteratively merge a sequence of proposed solutions into one 
optimal labeling. While similar to $\alpha$-expansion \cite{YURI_ALPHA_PAMI04},
the main difference is that proposals do not have to be constant labelings.

First, the initial labeling is computed: each blob is either assigned to the outlier model or to the first model in the pool based on the data term \eqref{eq:Error function geom and likelihood}. Then, at each iteration, the current solution $ l^0$ is ``fused'' with some new labeling proposal $l^1$ that was also formed by combining a model from the pool and the outlier model.

\begin{algorithm} 
\DontPrintSemicolon 
\SetKwInOut{Input}{Input}  
\Input{$blobs$ {//} text candidates\; \\
$models$ {//} text lines 
}
\SetKwInOut{Output}{Output}  
\Output{$labeling$ {//} blob-to-model map  
}
$l_0 \gets MakeLabeling(\emptyset, pool[0],blobs);$\;

\For{$i\leftarrow 1$ \KwTo $n\_models$} 
{ 
$l^1 \gets MakeLabeling(\emptyset, pool[i],blobs);$\;
$l^0 \gets FuseLabeling(l^0, l^1);$\;
}
$labeling   \gets l^0$\;
\Return{$labeling;$}\;
\caption{{\sc AssignModels} }
\label{algo:AssignModels}
\end{algorithm}

The {\em fusion move} is a binary optimization procedure for merging two solutions $l^1$ and $ l^0$ combining their best properties in one 
new labeling $\boldsymbol{l}$. 
This is analogous to crossover in genetic algorithms. There are many different 
ways to fuse two solutions. Any fusion result $\boldsymbol{l}$ is uniquely defined 
by a binary vector $\boldsymbol{x}$ specifying which of the components 
are borrowed from solution $l^1$ and which are from $ l^0$. In particular,
\begin{align}
\boldsymbol{l}(\boldsymbol{x}) = \boldsymbol{x}\circ l^1 + (1-\boldsymbol{x})\circ l^0,
\label{eq:optimal_l} 
\end{align}
where $\circ$ represents a point-wise product of two vectors, 
a.k.a. {\em Hadamard} product. 
In other words, each component of a fused labeling is defined by binary variable
$x_i\in\{0,1\}$ as $l_i = x_i\cdot l^1_i + (1-x_i)\cdot l^0_i$ for every 
text candidate $i\in \mathcal{I}$.

The goal of a {\em fusion move} is to find an optimal $\boldsymbol{x}$ giving the best possible crossover solution $\boldsymbol{l}(\boldsymbol{x})$ with respect to energy \eqref{eq:FusionFULLFUSION}. When model parameters $\boldsymbol{\theta}$ 
are fixed, \eqref{eq:FusionFULLFUSION} defines the energy for $\boldsymbol{l}(\boldsymbol{x})$ as a function of $\boldsymbol{x}$
\begin{align}
  E(\boldsymbol{l}(\boldsymbol{x}))   =& \sum\limits_{i\in \mathcal{I}}{D^0_i+(D^1_i-D^0_i)x_i} +\nonumber \\ 
 &\sum\limits_{j\in \mathcal{L}}{C_j \cdot (1-\mathrm{X}_{P_j^0}\mathrm{\overline{X}}_{P_j^1}) } +\nonumber \\ 
 &\sum\limits_{v\in \mathcal{V}}{C_v \cdot (1-\mathrm{X}_{P_v^0}\mathrm{\overline{X}}_{P_v^1}) } ,
\label{eq:FusionCutEnergy} 
\end{align}
where $ \mathcal{I} $ is the set of all detected text candidates, $ \mathcal{L} $ is the set of proposed text line models, $ \mathcal{V} $ is the set of languages. The data term is defined as
\begin{align}
D^0_i = D_i (l^0_i | \boldsymbol{\theta}) \\ \nonumber
D^1_i = D_i (l^1_i | \boldsymbol{\theta}).
\end{align}
Constants $ C_j $ and $ C_v $ are the costs of each text line model 
and each language as in energy \eqref{eq:FusionFULLFUSION}, and
\begin{align}
\label{eq:XX_switch} 
X_{P^0_j} = \prod\limits_{p \in P^0_j}{x_p}  \\ 
\nonumber
\overline{X}_{P^1_j} = \prod\limits_{p \in P^1_j}{\bar{x}_p},
\end{align}
where $\bar{x}_p = 1-x_p$ and $P^k_j = \{i\in\mathcal{I} | l^k_i = j\}$. 

Note that $\mathrm{X}_{P_j^0}\mathrm{\overline{X}}_{P_j^1} $ is a boolean expression that turns ``on'' and ``off'' text line costs in \eqref{eq:FusionCutEnergy}. The switch is ``on'' if there is at least one text candidate that supports text line model $j$ and the cost of the model is paid. Otherwise, the switch is ``off'' and the penalty is not paid. $\mathrm{X}_{P_v^0}\mathrm{\overline{X}}_{P_v^1} $ is defined similarly to \eqref{eq:XX_switch}.

Binary energy \eqref{eq:FusionCutEnergy} includes
unary potentials and high-order terms that came from the hierarchical
label costs in energy \eqref{eq:FusionFULLFUSION}. Similar high-order
potentials were studied by Delong et al. \cite{Delong:nips12} in 
the context of hierarchical clustering. We find a globally
optimal {\em fusion move} minimizing \eqref{eq:FusionCutEnergy} using the
same graph cut construction as proposed in \cite{Delong:nips12}.

\section{Evaluation} \label{sec:eval}

Our database of signboards from the Seoul subway consist of 500 images taken with a Samsung Galaxy SII, Galaxy S, Apple iPhone 3GS, Canon EOS 450D. 400 images were used for training and the remaining 100 for testing of both the AdaBoost classifier and the whole algorithm evaluation. 

We want to test all three modules of the algorithm in an isolated fashion. However, it is not straightforward to check how well detected blobs cover text on an image before the text lines are extracted. Instead, we compare the final line segments detected in one image to two ground truths - the first containing blobs detected programmatically by the blob detection module (Artificial base), the second containing blobs placed by a human (Real base). The difference between the results of these two experiments, see Table \ref{tab:text_detection_accuracy}, reflects the quality of the blob detector.

\begin{figure*}
         \centering
         \begin{subfigure}[b]{0.5\textwidth}
                 \includegraphics[width=\textwidth]{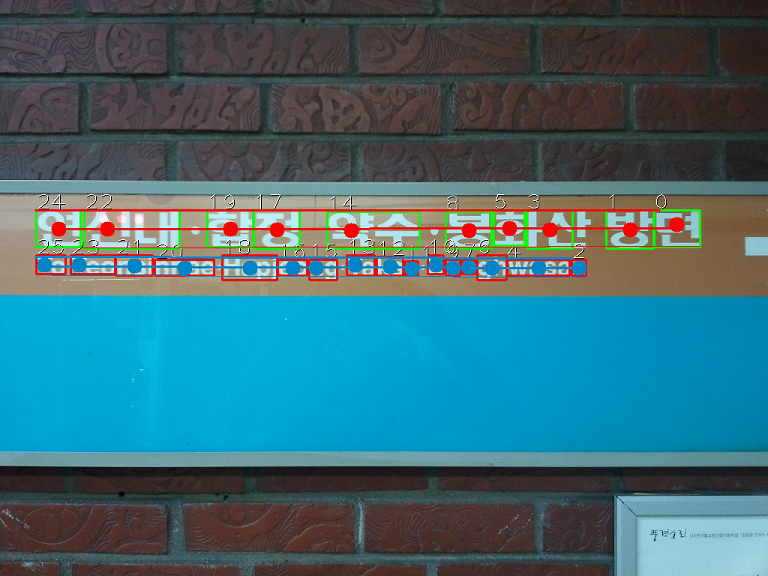}
                 \caption{}
         \end{subfigure}%
         ~ 
         \begin{subfigure}[b]{0.5\textwidth}
                 \includegraphics[width=\textwidth]{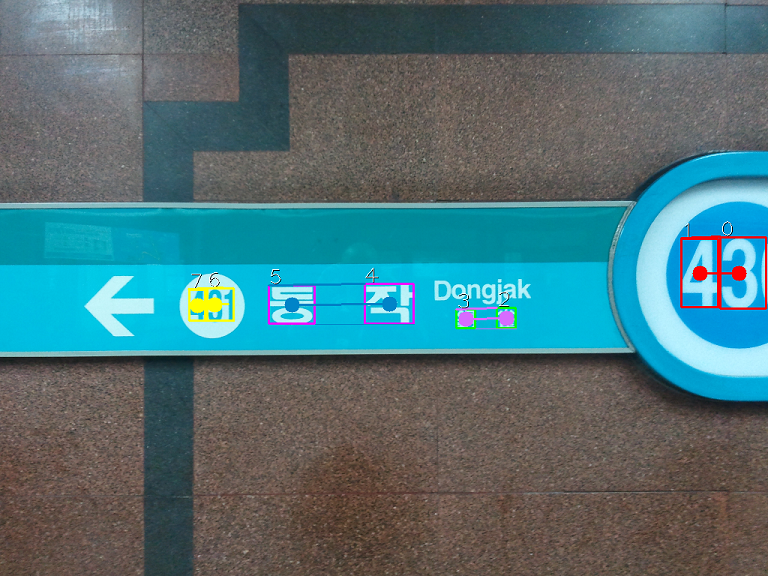}
                 \caption{}
         \end{subfigure}

         \begin{subfigure}[b]{0.5\textwidth}
                 \includegraphics[width=\textwidth]{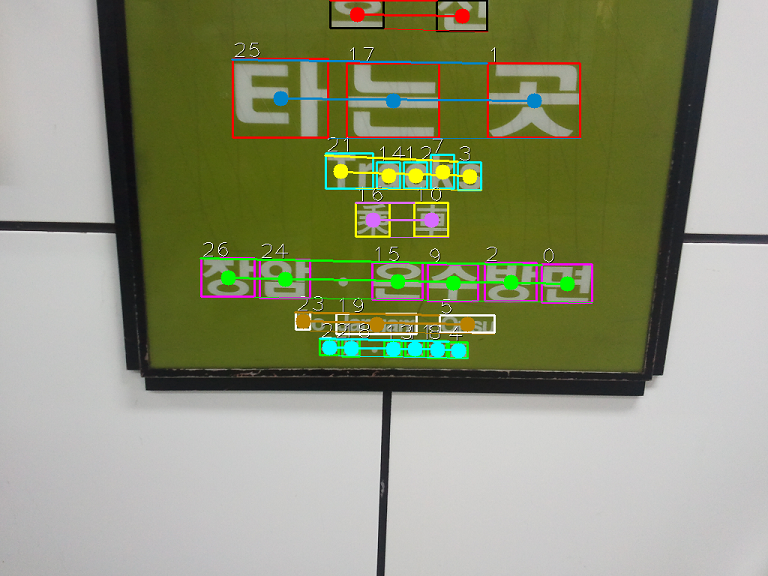}
                 \caption{}
         \end{subfigure}%
         ~ 
         \begin{subfigure}[b]{0.5\textwidth}
                 \includegraphics[width=\textwidth]{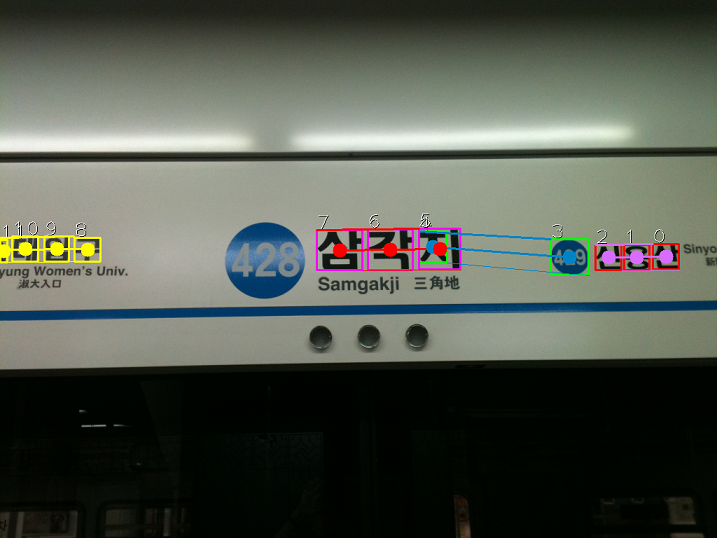}
                 \caption{}
         \end{subfigure}

         \begin{subfigure}[b]{0.5\textwidth}
                 \includegraphics[width=\textwidth]{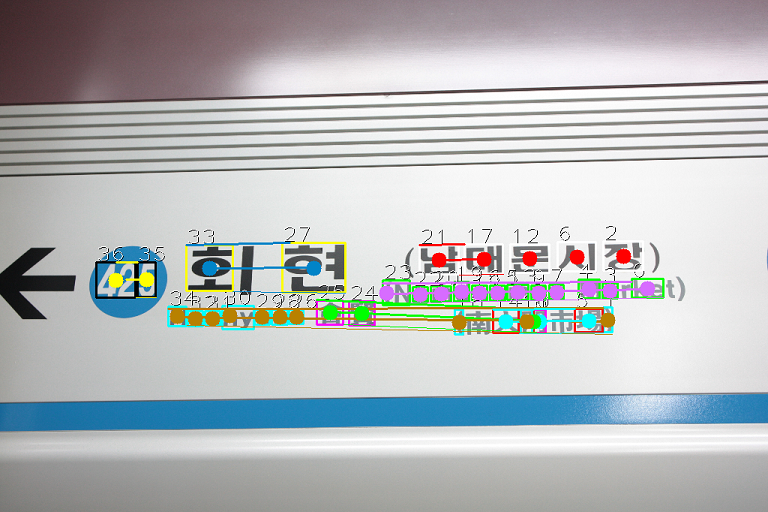}
                 \caption{}
         \end{subfigure}%
         ~ 
         \begin{subfigure}[b]{0.5\textwidth}
                 \includegraphics[width=\textwidth]{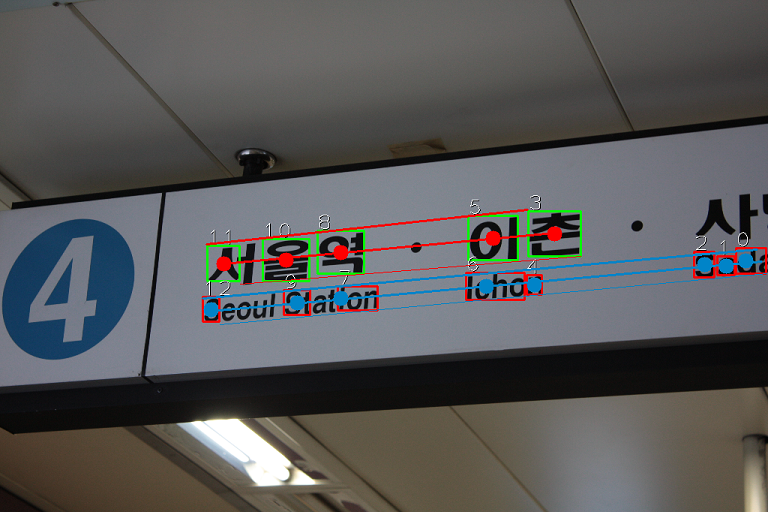}
                 \caption{}
         \end{subfigure}      
       
         \caption{Results of text line detection. Triplet of lines in one color shows one line. Circles mark centers of text  candidates. Blobs assigned to the outlier label are hidden for readability.}\label{fig:results}
\end{figure*}

\begin{table}[h]
\centering
\begin{tabular}{ |c|c|c|c| }
   \hline
   Ground truth  & Recall (\%) &  Precision (\%) &  F (\%) \\\hline\hline
   Real base   & 71   & 84   & 76 \\ \hline 
	Artificial base   & 76   & 84  & 79  \\	  
   \hline
\end{tabular} 
\caption{Text line detection accuracy.}
\label{tab:text_detection_accuracy}
\end{table}
 The confusion matrix for text candidate classification by AdaBoost is shown in Table ~\ref{tab:confusion_matrix}. The average recognition rate is 83.35 \%. The examples of text line detection is shown in Fig. \ref{fig:results}.
\begin{table}[h]
\centering
\begin{tabular}{|l||*{5}{c|}}\hline
\backslashbox{Actual}{Predicted}&
\rotatebox[origin=c]{90}{ English}&
\rotatebox[origin=c]{90}{ Korean}&
\rotatebox[origin=c]{90}{ Chinese }&
\rotatebox[origin=c]{90}{ Digit }&
\rotatebox[origin=c]{90}{ NonText }\\\hline\hline
English &902     &10             &16      &10      &45	\\\hline
Korean 	&52      &815            &63      &6       &40	\\\hline
Chinese &27      &14             &172     &0       &11	\\\hline
Digit 	&19      &1              &0       &121     &5	\\\hline
NonText &482     &305            &188     &144     &5278	\\\hline
\end{tabular}
\caption{Confusion matrix for classification.}
\label{tab:confusion_matrix} 
\end{table}

\section{Conclusions} 
In this work we introduce a challenging new problem of the multilingual multi-line text detection. We formulate the problem as a hierarchical MDL energy optimization and demonstrated that a fusion based method efficiently obtains good quality solutions for this energy. We obtain very promising results on our large database of images from the subway of metropolitan area of Seoul that we plan to make public for other researchers in computer vision.
Our experiments show that the bottleneck of our algorithm is the AdaBoost classifier performance, which could be enhanced. Extending our current hierarchy {\em blobs}$\rightarrow${\em lines}$\rightarrow${\em languages} into {\em blobs}$\rightarrow$ {\em characters} $\rightarrow${\em lines}$\rightarrow${\em languages} could further improve the results.
\bibliographystyle{splncs}
\bibliography{egbib}
\end{document}